\pdfoutput=1
\documentclass[10pt,twocolumn,letterpaper]{article}

\usepackage{cvpr}
\usepackage{times}
\usepackage{epsfig}
\usepackage{graphicx}
\usepackage{amsmath}
\usepackage{amssymb}

\usepackage[breaklinks=true,bookmarks=false]{hyperref}

\graphicspath{{images/}}
\usepackage{breqn}
\usepackage{multirow}
\usepackage{hhline}
\usepackage{ltablex}
\usepackage{siunitx}
\usepackage{caption}
\usepackage{booktabs}
\usepackage{mathtools}
\usepackage{arydshln}

\usepackage{capt-of}
\usepackage{float}
\usepackage{wrapfig}
\usepackage{bm}
\usepackage{afterpage}
\usepackage[export]{adjustbox}
\usepackage{amssymb,amsmath,amsthm,enumitem}

\cvprfinalcopy 

\ifcvprfinal\fi
\begin{document}
\title{Hierarchical Cross-Modal Talking Face Generation \\with Dynamic Pixel-Wise Loss}
\author{Lele Chen \quad Ross K. Maddox \quad Zhiyao Duan \quad Chenliang Xu\\
University of Rochester, USA \\
{\tt\small {\{lchen63, rmaddox\}}@ur.rochester.edu}, {\tt\small{\{zhiyao.duan, chenliang.xu\}}@rochester.edu}
}
\maketitle

\begin{abstract}
We devise a cascade GAN approach to generate talking face video, which is robust to different face shapes, view angles, facial characteristics, and noisy audio conditions. Instead of learning a direct mapping from audio to video frames, we propose first to transfer audio to high-level structure, i.e., the facial landmarks, and then to generate video frames conditioned on the landmarks. Compared to a direct audio-to-image approach, our cascade approach avoids fitting spurious correlations between audiovisual signals that are irrelevant to the speech content. We, humans, are sensitive to temporal discontinuities and subtle artifacts in video. To avoid those pixel jittering problems and to enforce the network to focus on audiovisual-correlated regions, we propose a novel dynamically adjustable pixel-wise loss with an attention mechanism. Furthermore, to generate a sharper image with well-synchronized facial movements, we propose a novel regression-based discriminator structure, which considers sequence-level information along with frame-level information. Thoughtful experiments on several datasets and real-world samples demonstrate significantly better results obtained by our method than the state-of-the-art methods in both quantitative and qualitative comparisons.
\end{abstract}

\section{Introduction}
\label{sec:intro}

\begin{figure}[t]
\centering
\includegraphics[width=0.95\linewidth]{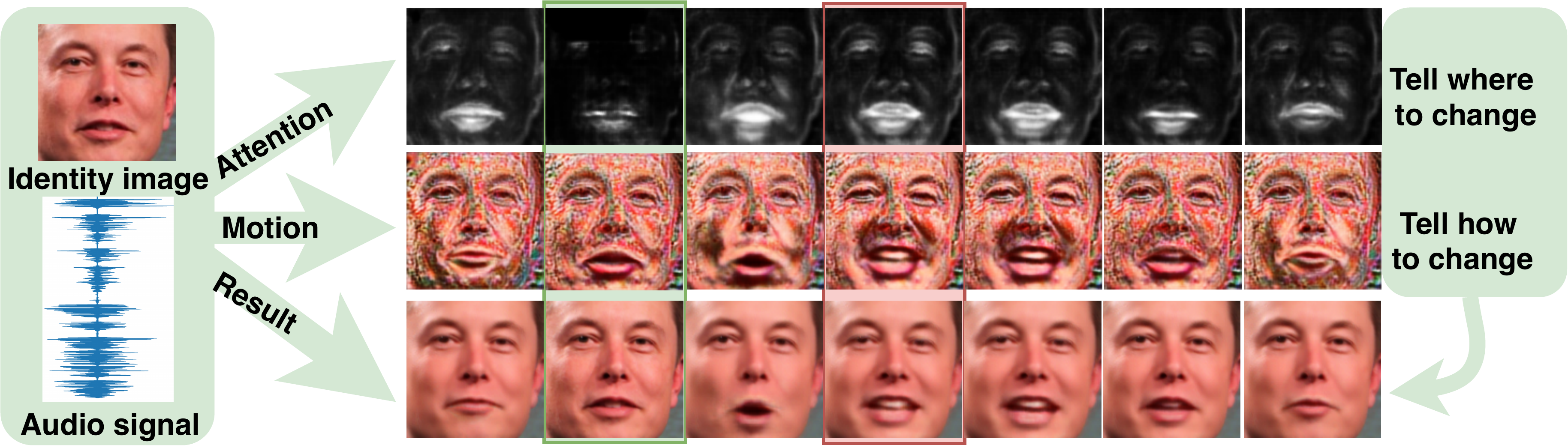}
\caption{Problem description. The model takes an arbitrary audio speech and one face image, and synthesizes a talking face saying the speech. The synthesized frames (last row) consist of synthesized attention (first row) and motion (second row), which demonstrate where and how the dynamics are synthesizing. For example, the face in the green box looks similar to the example face so that the attention map is almost dark; the face in the red box differs much from the example image, and hence the attention highlights the mouth region and the motion part hints white pixels for teeth.}
\label{fig:teaser}
\end{figure}

Modeling the dynamics of a moving human face/body conditioned on another modality is a fundamental problem in computer vision, where applications are ranging from audio-to-video generation~\cite{obamanet,ChungJZ17,ChenSDX17} to text-to-video generation~\cite{MarwahMB17,LiMSCC18} and to skeleton-to-image/video generation~\cite{MaJSSTG17,di2018gp}. This paper considers such a task: given a target face image and an arbitrary speech audio recording, generating a photo-realistic talking face of the target subject saying that speech with natural lip synchronization while maintaining a smooth transition of facial images over time (see Fig.~\ref{fig:teaser}). Note that the model should have a robust generalization capability to different types of faces (e.g., cartoon faces, animal faces) and to noisy speech conditions  (see Fig.~\ref{fig:visual}). Solving this task is crucial to enabling many applications, e.g., lip-reading from over-the-phone audio for hearing-impaired people, generating virtual characters with synchronized facial movements to speech audio for movies and games.

The main difference between still image generation and video generation is temporal-dependency modeling. There are two main reasons why it imposes additional challenges: people are sensitive to any pixel jittering (e.g., temporal discontinuities and subtle artifacts) in a video; they are also sensitive to slight misalignment between facial movements and speech audio. However, recent researchers~\cite{ChungJZ17,zhou2018talking,DBLP:journals/tog/KarrasALHL17} tended to formulate video generation as a temporally independent image generation problem. For example, Chung et al.~\cite{ChungJZ17} proposed an encoder-decoder structure to generate one image from 0.35-second audio at each time. Song et al.~\cite{yangsong} adopted a recurrent network to consider temporal dependency. They applied RNN in the feature extraction part, however, each frame was generated independently in the generation stage. In this paper, we propose a novel temporal GAN structure, which consists of a multi-modal convolutional-RNN-based (MMCRNN) generator and a novel regression-based discriminator structure. By modeling temporal dependencies, our MMCRNN-based generator yields smoother transactions between adjacent frames. Our regression-based discriminator structure combines sequence-level (temporal) information and frame-level (pixel variations) information to evaluate the generated video. 

Another challenge of the talking face generation is to handle various visual dynamics (e.g., camera angles, head movements) that are not relevant to and hence cannot be inferred from speech audio. Those complicated dynamics, if modeled in the pixel space~\cite{villegas17mcnet}, will result in low-quality videos. For example, in web videos~\cite{DBLP:conf/accv/ChungZ16a,Nagrani17} (e.g., LRW and VoxCeleb datasets), speakers move significantly when they are talking. Nonetheless, all the recent photo-realistic talking face generation methods~\cite{ChungJZ17,zhou2018talking,yangsong,ChenLMDX18,obamanet,DBLP:conf/eccv/WilesKZ18} failed to consider this problem. In this paper, we propose a hierarchical structure that utilizes a high-level facial landmarks representation to bridge the audio signal with the pixel image. Concretely, our algorithm first estimates facial landmarks from the input audio signal and then generates pixel variations in image space conditioned on generated landmarks. Besides leveraging intermediate landmarks for avoiding directly correlating speech audio with irrelevant visual dynamics, we also propose a novel dynamically adjustable loss along with an attention mechanism to enforce the network to focus on audiovisual-correlated regions. It is worth to mention that in a recent audio-driven facial landmarks generation work~\cite{DBLP:conf/ica/EskimezMXD18}, such irrelevant visual dynamics are removed in the training process by normalizing and identity-removing the facial landmarks. This has been shown to result in more natural synchronization between generated mouth shapes and speech audio.

Combining the above features, which are designed to overcome limitations of existing methods, our final model can capture informative audiovisual cues such as the lip movements and cheek movements while generating robust talking faces under significant head movements and noisy audio conditions. We evaluate our model along with state-of-the-art methods on several popular datasets (e.g., GRID~\cite{Cooke2006}, LRW~\cite{DBLP:conf/accv/ChungZ16a}, VoxCeleb~\cite{Nagrani17} and TCD~\cite{DBLP:journals/tmm/HarteG15}). Experimental results show that our model outperforms all compared methods and all the proposed features contribute effectively to our final model. Furthermore, we also show additional novel examples of synthesized facial movements of the human/cartoon characters who are not in any dataset to demonstrate the robustness of our approach. 

The contributions of our work can be summarized as follows: (1) We propose a novel cascade network structure to reduce the effects of the sound-irrelevant visual dynamics in the image space. Our model explicitly constructs high-level representation from the audio signal and guides video generation using the inferred representation. (2) We exploit a dynamically adjustable pixel-wise loss along with an attention mechanism which can alleviate temporal discontinuities and subtle artifacts in video generation. (3) We propose a novel regression-based discriminator to improve the audio-visual synchronization and to smooth the facial movement transition while generating realistic looking images. The code has been released at \href{https://github.com/lelechen63/ATVGnet}{\bf https://github.com/lelechen63/ATVGnet}. 
\section{Related Work}
\label{sec:related}
In this section, we first briefly survey related work on the talking face generation task. Then we discuss the related work of each technique used in our model.

\noindent \textbf{Talking Face Synthesizing} \quad The success of traditional approaches has been mainly limited to synthesizing a talking face from speech audio of a specific person~\cite{DBLP:journals/cgf/GarridoVSSVPT15,DBLP:conf/icassp/FanWSX15,DBLP:journals/tog/SuwajanakornSK17}. For example, Suwajanakorn et al.~\cite{DBLP:journals/tog/SuwajanakornSK17} synthesized a taking face of President Obama with accurate lip synchronization, given his speech audio. The mechanism is to first retrieve the best-matched lip region image from a database through audiovisual feature correlation and then compose the retrieved lip region with the original face. However, this method requires a large amount of video footage of the target person. More recently, by combining the GAN/encoder-decoder structure and the data-driven training strategy,~\cite{yangsong,ChungZ16,ChenLMDX18,zhou2018talking} can generate arbitrary faces from arbitrary input audio.

\noindent \textbf{High-Level Representations}\quad In recent years, high-level representations of  images~\cite{VillegasYZSLL17,NIPS2018_7536,WichersVEL18,Hong_2018_CVPR} have been exploited in video generation tasks by using an encoder-decoder structure as the main approach. Given a condition, we can transfer it to high-level representations and feed them to a generative network to output a distribution over locations that a pixel is predicted to move. By adopting human body landmarks, Villegas et al.~\cite{VillegasYZSLL17} proposed an encoder-decoder network which achieves long-term future prediction. Suwajanakorn et al.~\cite{obamanet} transferred the audio signal to lip shapes and then synthesized the mouth texture based on the transferred lip shapes. These works have inspired us to use the facial landmarks to bridge audio with row pixel generation. 
\begin{figure*}[t]
\centering
\includegraphics[width=\linewidth]{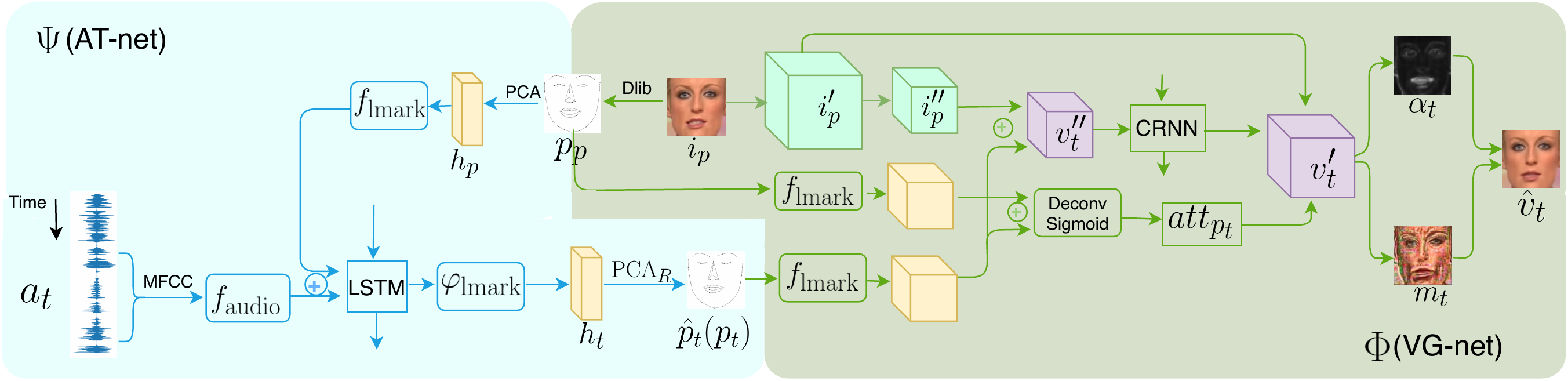}
\caption{Overview of our network architecture. The blue part illustrates the AT-net, which transfers audio signal to low-dimensional landmarks representation and the green part illustrates the VG-net, which generates video frames conditioned on the landmark. During training, the input to VG-net are ground truth landmarks ($p_{1 \colon T}$). During inference, the input to VG-net are fake landmarks ($\hat{p}_{1 \colon T}$) generated by AT-net. The AT-net and VG-net are trained separately to avoid error accumulation.}
\label{fig:overall}
\end{figure*}

\noindent \textbf{Attention Mechanism} \quad Attention mechanism is an emerging topic in natural language tasks~\cite{DBLP:conf/emnlp/LuongPM15} and image/video generation task~\cite{DBLP:conf/eccv/PumarolaAMSM18,DBLP:journals/corr/abs-1805-08318,Ma_2018_CVPR,Xu_2018_CVPR}. Pumarola et al.~\cite{DBLP:conf/eccv/PumarolaAMSM18} generated facial expression conditioned on action units annotations. Instead of using a basic GAN structure, they exploited a generator that regresses an attention mask and a RGB color transformation over the entire image. The attention mask defines a per-pixel intensity specifying to what extend each pixel of the original image will contribute to the final rendered image. We adopt this attention mechanism to make our network robust to visual variations and noisy audio conditions. Feng et al.~\cite{FengWSWZ18} observed that integrating a weighted mask into the loss function during training can improve the performance of the reconstruction network. Based on this observation, rather than using a fixed loss weights, we propose a dynamically adjustable loss by leveraging the attention mechanism to emphasize the audiovisual regions.
\section{Architecture}
\label{sec:model}
This section describes the architecture of the proposed model. Fig.~\ref{fig:overall} shows the overall diagram, which is decoupled into two parts: audio transformation network (AT-net) and visual generation network (VG-net). First, we explain the overall architecture and the training strategy in Sec.~\ref{subsec:decouple}. Then, we introduce two novel components: attention-based dynamic pixel-wise loss in Sec.~\ref{subsec:attention} and a regression-based discriminator structure in Sec.~\ref{subsec:discrimiator} used in our VG-net. 
\subsection{Overview}
\label{subsec:decouple}
\noindent \textbf{Cascade Structure and Training Strategy} \quad We tackle the task of talking face video generation in a cascade perspective. Given the input audio sequence $a_{1 \colon T}$, one example frame $i_p$ and its landmarks $p_p$, our model generates facial landmarks sequence $\hat{p}_{1 \colon T}$ and subsequently generates frames $\hat{v}_{1 \colon T}$. To solve this problem, we come up with a novel cascade network structure: 
\begin{align}
\label{eq:train_psi}
\hat{p}_{1 \colon T} &= \Psi(a_{1 \colon T},p_{p}) \enspace,\\ 
\label{eq:train_phi}
\hat{v}_{1 \colon T} &= \Phi(\hat{p}_{1 \colon T}, i_{p}, p_p) \enspace,
\end{align}
where the AT-net $\Psi$ (see Fig.~\ref{fig:overall} blue part) is a conditional LSTM encoder-decoder and the VG-net $\Phi$ (see Fig.~\ref{fig:overall} green part) is a multi-modal convolutional recurrent network. During inference, the AT-net $\Psi$ (see Eq.~\ref{eq:train_psi}) observes audio sequence $a_{1 \colon T} $ and example landmarks $p_p $ and then predicts low-dimensional facial landmarks $\hat{p}_{1 \colon T}$. By passing $\hat{p}_{1 \colon T}$ into VG-net $\Phi$ (see Eq.~\ref{eq:train_phi}) along with example image $i_p$ and $p_p$, we subsequently get synthesized video frames $\hat{v}_{1 \colon T}$. $\Psi$ and $\Phi$ are trained in a decoupled way so that $\Phi$ can be trained with teacher forcing strategy. To avoid the error accumulation caused by $\hat{p}_{1 \colon T}$, $\Phi$ is conditioned on ground truth landmarks $p_{1 \colon T}$ during training. 

\noindent \textbf{Audio Transformation Network (AT-net)} \quad Specifically, the AT-net ($\Psi$) is formulated as:
\begin{align}
\label{eq:at}
[h_{t},c_{t}]&=\varphi_{\text{lmark}}(\text{LSTM}(f_{\text{audio}}(a_{t}),f_{\text{lmark}}( h_{p}), c_{t-1})),\\
\label{eq:pca}
\hat{p}_{t} &=\text{PCA}_{\text{R}}(h_{ t}) = h_{t} \odot \omega * \text{U}^T + \text{M}
\enspace. 
\end{align}
Here, the AT-net observes the audio MFCC $a_{t}$ and landmarks PCA components $h_{p}$ of the target identity and outputs PCA components $h_t$ that are paired with the input audio MFCC. The $f_{\text{audio}}, f_{{\text{lmark}}}$ and $\varphi_{\text{lmark}}$ indicate audio encoder, landmarks encoder and landmarks decoder. The $c_{t-1}$ and $c_t$ are outputs from cell units. $\text{PCA}_{\text{R}}$ is PCA reconstruction and $\omega$ is a boost matrix to enhance the PCA feature. The $\text{U}$ corresponds to the largest eigenvalues and $\text{M}$ is the mean shape of landmarks in the training set. In our empirical study, we observe that PCA can decrease the effect of none-audio-correlated factors (e.g., head movements) for training the AT-net.

\noindent \textbf{Visual Generation Network (VG-net)} \quad  Intuitively, similar to~\cite{WichersVEL18,VillegasYZSLL17}, we assume that the distance between current landmarks $p_{t}$ and example landmarks $p_{p}$ in feature space can represent the distance between current image frame and example image in image feature space. Based on this assumption (see Eq.~\ref{eq:generator0}), we can obtain current frame feature $v''_t$ (size of $128 \times 8 \times 8$). Different from their methods, we replace element-wise addition with channel-wise concatenation in Eq.~\ref{eq:generator0}, which better preserves original frame information in our empirical study. At the meanwhile, we can also compute an attention map ($att_{p_t}$) based on the difference between $p_t$ and $p_p$ (see Eq.~\ref{eq:generator2}). By feeding the computed $v''_t$ and $att_{p_t}$ along  with example image feature $i'_p$ (size of $128 \times 32 \times 32$) into the MMCRNN part, we obtain the current image feature $v'_t$ (see Eq.~\ref{eq:generator3}). The resultant image feature $v'_t$ will be used to generate video frames as detailed in the next section. Specifically, the VG-net is performed by:
\begin{align}
\label{eq:generator0}
&v''_t = f_{{\text{img}}(i_p)} \oplus( f_{{\text{lmark}}}(p_{t}) - f_{{\text{lmark}}}(p_p))
\enspace, \\
\label{eq:generator2}
&att_{p_t} = \sigma(f_{{\text{lmark}}}(p_t) \oplus f_{{\text{lmark}}}(p_p))
\enspace, \\
\label{eq:generator3}
&v'_t = ({\text{CRNN}}( v''_t)) \odot {\text{att}}_{p_t} + i'_p  \odot (\textbf{1} - {\text{att}}_{p_t}) \enspace , &&
\end{align}
where $\oplus$ and $\odot$ are concatenation operation and element-wise multiplication, respectively. The CRNN part consists of Conv-RNN, residual block and deconvolution layers. $i'_p$ is the middle layer output of $f_{{\text{img}}}(i_p)$, and $\sigma$ is Sigmoid activation function. We omit some convolution operations in equations for better understanding. 
\begin{figure}[t]
\centering
\includegraphics[width=0.9\linewidth]{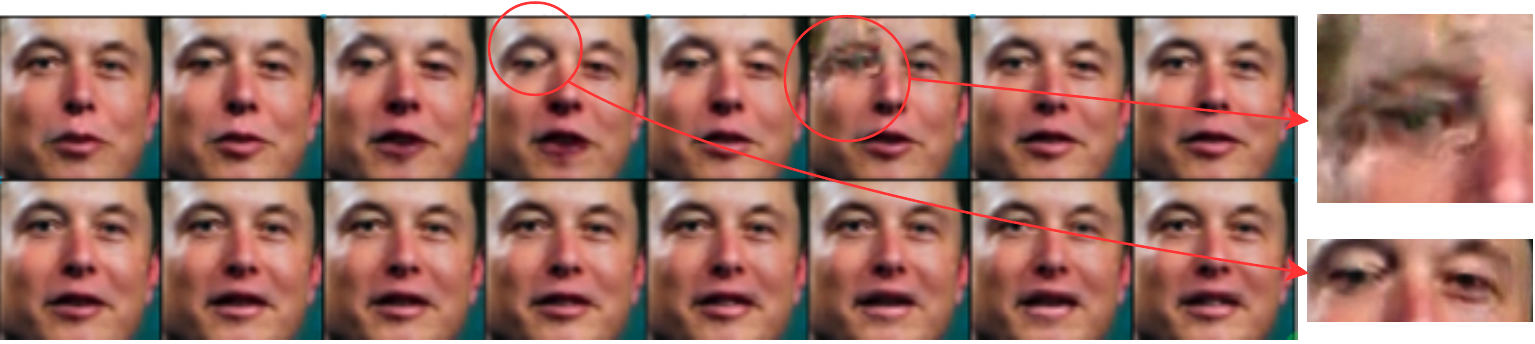}
\caption{The results of our baseline method. The synthesized frames with pixel jittering problem. The discontinuous problem and subtle artifacts will be amplified after composing into a video.}
\label{fig:pixeljettering}
\end{figure}
\subsection{Attention-Based Dynamic Pixel-wise Loss}
\label{subsec:attention}
Recent works on video generation adopt either GAN-based methods~\cite{ChenLMDX18,VondrickPT16,yangsong} or Encoder-Decoder-based methods~\cite{ChungJZ17}. However, one common problem is the pixel jittering between adjacent frames (see Fig.~\ref{fig:pixeljettering}). Pixel jittering is not obvious in single image generation but is a severe problem for video generation as humans are sensitive to any pixel jittering, e.g., temporal discontinuities and subtle artifacts in a video. The reason is that GAN loss or L1/L2 loss can barely generate perfect frames that all pixels are consistently changing in temporal domain, especially for audiovisual-non-correlated regions, e.g., background and head movements. In order to solve the pixel jittering problem, we propose a novel dynamic pixel-wise loss to enforce the generator to generate consistent pixels along temporal axis. 

As mentioned in Sec.~\ref{sec:related}, Pumarola et al.~\cite{DBLP:conf/eccv/PumarolaAMSM18} exploited a generator that regresses an attention mask and a RGB color transformation over the entire image. We adapt this attention mechanism in our VG-net to disentangle the motion part from audiovisual-non-correlated regions. Therefore, our final frame output is governed by the combination:
\begin{align}
\label{eq:motion_att}
\hat{v}_{t} =  \bm{\alpha}_{t} \odot m_t + (\textbf{1} - \bm{\alpha}_{t}) \odot i_{p}
\enspace,
\end{align}
where attention ${\bm{\alpha}}_{t}$ is obtained by applying convolution and Sigmoid activation operations on $v'_t$, motion $m_{t}$ is obtained by applying another convolution and hyperbolic tangent activation operations on $v'_t$. This step enforces the network to generate stable pixels in audiovisual-non-correlated regions while generating movements in audiovisual-correlated regions.

From Fig.~\ref{fig:result1}, we can conclude that the pixels in audiovisual-non-correlated regions (e.g., hair, background etc.) usually attract less attention and are irrelevant to given condition (audio). In contrast, the network is mainly focusing on correlated regions (e.g., mouth, jaw, and cheek). Intuitively, $\bm{0} \leq {\bm{\alpha}}_{t} \leq \bm{1} $ can be viewed as a spatial mask that indicates which pixels of given face image $i_p$ need to move at time step $t$. We can also regard ${\bm{\alpha}}_{t}$ as a reference to represent to what extend each pixel contributes to the loss. The audiovisual-non-correlated regions should contribute less to the loss compared with the correlated regions. Thus, we propose a novel dynamic adjustable pixel-wise loss by leveraging the power of $\bm{\alpha}_{t}$, which is defined as: 
\begin{align}
\label{eq:l1loss}
\mathcal{L}_{{\text{pix}}} = \sum\limits_{t=1}^T\lVert{(v_{t} - \hat{v_t}) \odot (\overline{\bm{\alpha}}_t + \bm{\beta})}\rVert_{1})
\enspace,
\end{align}
where $\overline{\bm{\alpha}}_{t}$ is the same as ${\bm{\alpha}}_{t}$ but without gradient. It represents the weight of each pixel dynamically that eases the generation. We remove the gradient of $\bm{\alpha}_t$ when back-propagating the loss to the network to prevent trivial solutions (lower loss but no discriminative ability). We also give base weights $\bm{\beta}$ to all pixels to make sure all pixels will be optimized. Here, we manually tune the hyper-parameter $\bm{\beta}$ and set $\bm{\beta}= {\bf{0.5}}$ in all of our experiments.
\begin{figure}[t]
\centering
\includegraphics[width=0.85\linewidth]{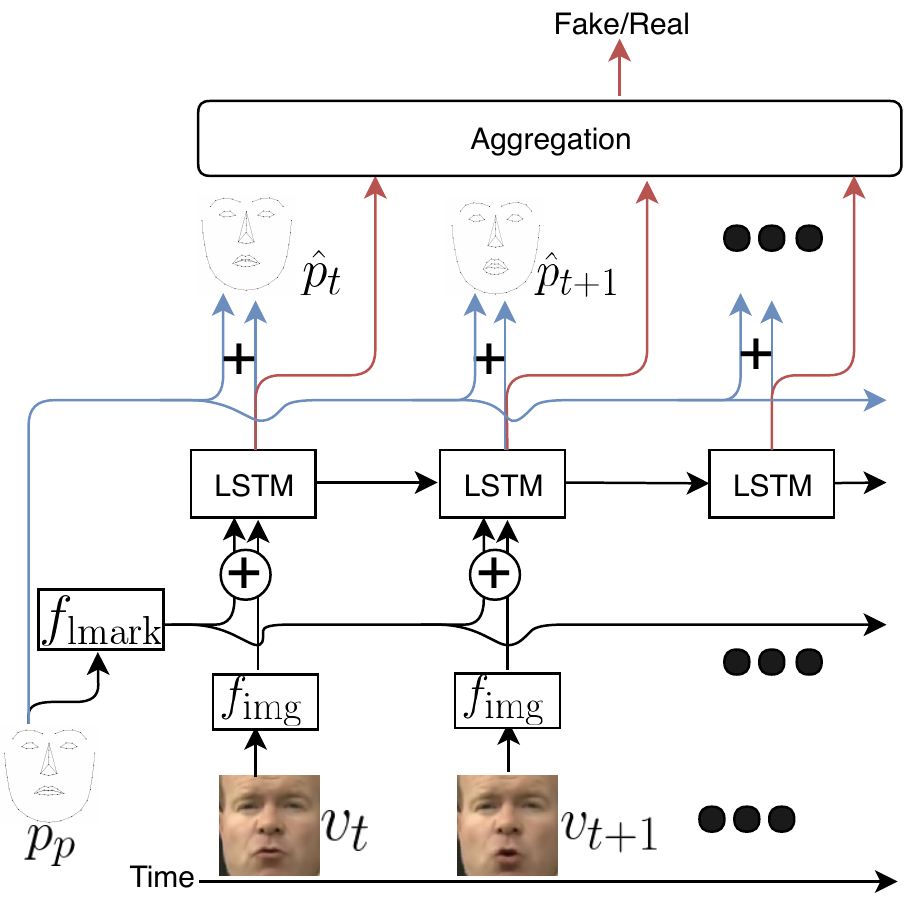}
\caption{The overview of the regression-based discriminator. The $\oplus$ means concatenation. The $+$ means element-wise addition. The blue arrow and red arrow represent $\text{D}_p$ and $\text{D}_s$, respectively.}
\label{fig:disc}
\end{figure}
\subsection{Regression-Based Discriminator}
\label{subsec:discrimiator}
Recently, people find that perceptual loss~\cite{DBLP:conf/eccv/JohnsonAF16} is helpful for generating sharp images in GAN/VAE~\cite{yangsong,ChenLMDX18}. Perceptual loss utilizes high-level features to compare generated images and ground-truth images resulting in better sharpness of the synthesized images. The key idea is that the weights of the perceptual network part are fixed, and the loss will only contribute to the generator/decoder part. Based on this intuition, we propose a novel discriminator structure (see Fig.~\ref{fig:disc}). The discriminator observes example landmarks $p_p$ and either ground truth video frames $v_{1 \colon T}$ or synthesized video frames $\hat{v}_{1 \colon T}$, then regresses landmarks shapes $\hat{p}_{1 \colon T}$ paired with the input frames, and additionally, gives a discriminative score $s$ for the entire sequence. Specifically, we formulate discriminator into frame-wise part $\text{D}_p$ (blue arrows in Fig.~\ref{fig:disc}) and sequence-level part $\text{D}_s$ (red arrows in Fig.~\ref{fig:disc}).

The $\text{D}_p$ observes example landmarks and video frames, then regresses the landmarks sequence based on observed information. By yielding the facial landmark, it can evaluate the input image based on high-level representation in a frame-wise fashion. Specifically, the $\hat{p}_{t}$ is calculated by:
\begin{align}
\label{eq:pose_d}
\hat{p}_{t} & =\text{D}_p(p_p, v_{t})  \nonumber \\
&= p_p + {\text{LSTM}}(f_{{\text{lmark}}}(p_p) \oplus f_{{\text{img}}}(v_{t})) \enspace,
\end{align}
which observes ground truth image during discriminator training stage and observes synthesized image during generator training stage.

Besides $\text{D}_p$, the LSTM cell unit yields another branch $\text{D}_s$, which obtains vectors from each LSTM cell unit and aggregates them by \text{average pooling}. By passing through a Sigmoid activation function, $\text{D}_s$ yields final discriminative score $s$ for the overall input sequence. The score $s$ can obtained by:
\begin{align}
\label{eq:ds}
s &= \text{D}_s(p_p, v_{1 \colon T}) \nonumber \\
&=\sigma( \frac{1}{T}\sum_{t=1}^{T}({\text{LSTM}}(f_{{\text{lmark}}}(p_p) \oplus f_{{\text{img}}}(v_t)))) 
\enspace.
\end{align}
The $\text{D}_p$ part is optimized to minimize the L2 loss between the predicted landmarks and the ground truth landmarks. Thus our GAN loss can be expressed as:
\begin{align}
\label{eq:dloss}
\mathcal{L}_{\text{gan}}= &\mathbb{E}_{p_p, v_{1 \colon T}}[\log {\text{D}_s}(p_p,v_{1 \colon T})] + \nonumber \\
&\mathbb{E}_{ p_p,p_{1 \colon T}, i_p}[\log (1- {\text{D}_s}(p_p, {\text{G}}(p_p,p_{1 \colon T}, i_p))] + \nonumber \\
   &  \lVert{({\text{D}_p}(p_p, {\text{G}}(p_p,p_{1 \colon T}, i_p)) - p_{1\colon T}) \odot {\text{M}}_{p}}\rVert^2_{2} + \nonumber \\
   &  \lVert{({\text{D}_p}(p_p,v_{1 \colon T}) - p_{1\colon T}) \odot {\text{M}}_{p}}\rVert^2_{2}
 \enspace ,
\end{align}
where $\text{M}_{p}$ is a pre-defined weight mask hyper-parameter which can penalize more on lip regions. By updating the parameters based on the regression loss when training the discriminator, the $\text{D}_p$ can learn to extract low-dimensional representations from raw image data. When we train the generator, we will fix the weights of discriminator including $\text{D}_s $ and $\text{D}_p $ so that $\text{D}_p $ will not compromise to generator. The loss back-propagated from $\text{D}_p$ will enforce generator to generate accurate face shapes (e.g., cheek shape, lip shape etc.) and the loss back-propagated from $ \text{D}_s$ will enforce the network to generate high-quality images.
\subsection{Objective Function}
\label{subsec:Objective}
By linearly combining all partial losses introduced in Sec.~\ref{subsec:attention} and Sec.~\ref{subsec:discrimiator}, the \textit{full loss} function $\mathcal{L}$ can be expressed as:
\begin{align}
\label{eq:full}
\mathcal{L}=   \mathcal{L}_{{\text{gan}}} + \lambda * \mathcal{L}_{{\text{pix}}}
 \enspace ,
\end{align}
where $\lambda$ is a hyper-parameter that controls the relative importance of different loss terms. We set $\lambda= 10.0 $ in our experiments.
\begin{figure*}[t]
\centering
\includegraphics[width=\linewidth]{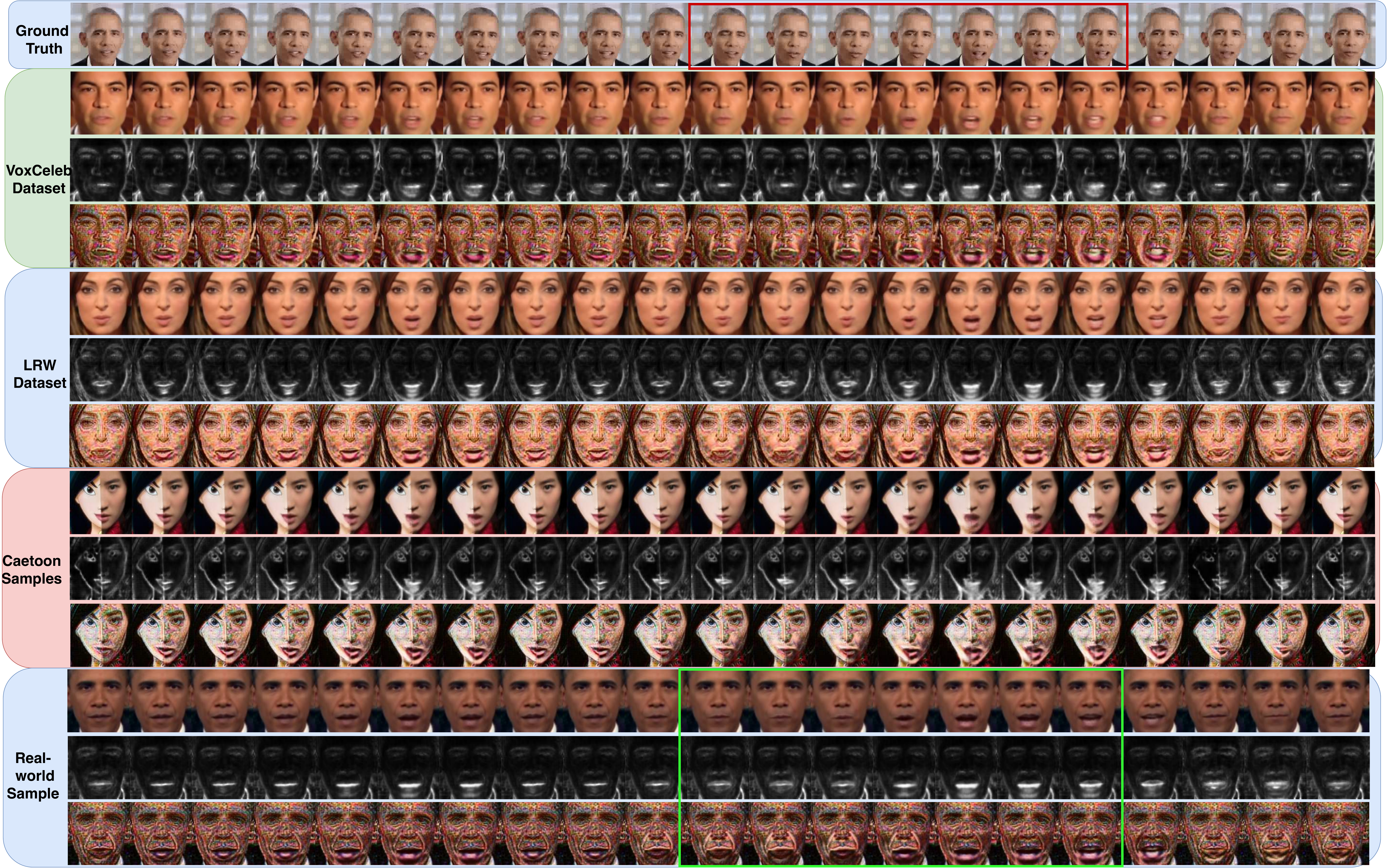}
\caption{The outputs of ATVGnet. The inputs are one real-world audio sequence and different example identity images range from real-world people to cartoon characters. The first row is ground truth images paired with the given audio sequence. We mark the different sources of the identity image on the left side. From this figure, we can find that the lip movements of our synthesized frames (e.g., the green box in the last row) are well-synchronized with the ground truth (red box in first row). Meanwhile, the attention (middle row of the green box) accurately indicates where need to move and the motion (last row of the green box) indicates what the dynamics look like (e.g. white pixels for teeth and red pixels for lips).}
\label{fig:result1}
\end{figure*}
\section{Experiments}
\label{sec:experiments}
In this section, we conduct thoughtful experiments to demonstrate the efficiency and effectiveness of the proposed architecture for video generation. Sec.~\ref{subsec:setup} explains datasets and implementation in detail. Sec.~\ref{subsec:quantitative} shows our results along with other state-of-the-art methods. We show user studies and ablation study in Sec.\ref{subsec:userstudy} and  Sec.~\ref{subsec:ablation} respectively. 
\begin{table*}[t]
\begin{center}
\begin{tabular}{c c c c c c}
\toprule
\toprule
Method & Real time & ATVGnet(our) & Chung et al.\cite{ChungJZ17} & Zhou et al.\cite{zhou2018talking} & Wiles et al.\cite{DBLP:conf/eccv/WilesKZ18}\\
\hline
Inference time (FPS)& 30 & 34.53  & 19.10 & 10.00 & 10.53   \\
\hline
\bottomrule
\end{tabular}
\end{center}
\caption{The inference time of difference models. We use frame rate (FPS) to measure the time.}
\label{tab:time}
\end{table*}
\subsection{Experimental Setup}
\label{subsec:setup}
\noindent \textbf{Dataset} \quad  We quantitatively and qualitatively evaluate our ATVGnet on LRW dataset~\cite{ChungZ16} and GRID dataset~\cite{Cooke2006}. The LRW dataset consists of 500 different words spoken by hundreds of different speakers in the wild. We follow the same train-test split as in~\cite{ChungZ16}. In GRID dataset, there are 1000 short videos, each spoken by 33 different speakers in the experimental condition. For the image stream, all the talking faces in the videos are aligned based on key-points (eyes and nose) of the extracted landmarks using ~\cite{King09} at the sampling rate of $25$FPS, and then resize to $128 \times 128$. As for audio data, each audio segment corresponds to 280ms audio. We extract MFCC at the window size of 10ms and use center image frame as the paired image data. Similar to ~\cite{ChungJZ17,yangsong}, we remove the first coefficient from the original MFCC vector, and eventually yield a $28 \times 12$ MFCC feature for each audio chunk.

\noindent \textbf{Implementation Details} \quad Our network is implemented using Pytorch 0.4 library. We adopt Adam optimizer during training with the fixed learning rate of $2 \times 10 \textsuperscript{-4}$. We initialize all network layers using random normalization with mean=0.0, std=0.2. All models are trained and tested on a single NVIDIA GTX 1080Ti. During the training, the AT-net converges after 3 hours and the VG-net is stable after 24 hours. Table~\ref{tab:time} shows the generation time during inference stage. We can find that our inference time can achieve around $34.5$ frames per second (FPS), which is much faster than~\cite{WichersVEL18,zhou2018talking, ChungJZ17} and slightly faster than real time (30 FPS).

\subsection{Results}
\label{subsec:quantitative}

\begin{figure*}[t]
\centering
\includegraphics[width=\linewidth]{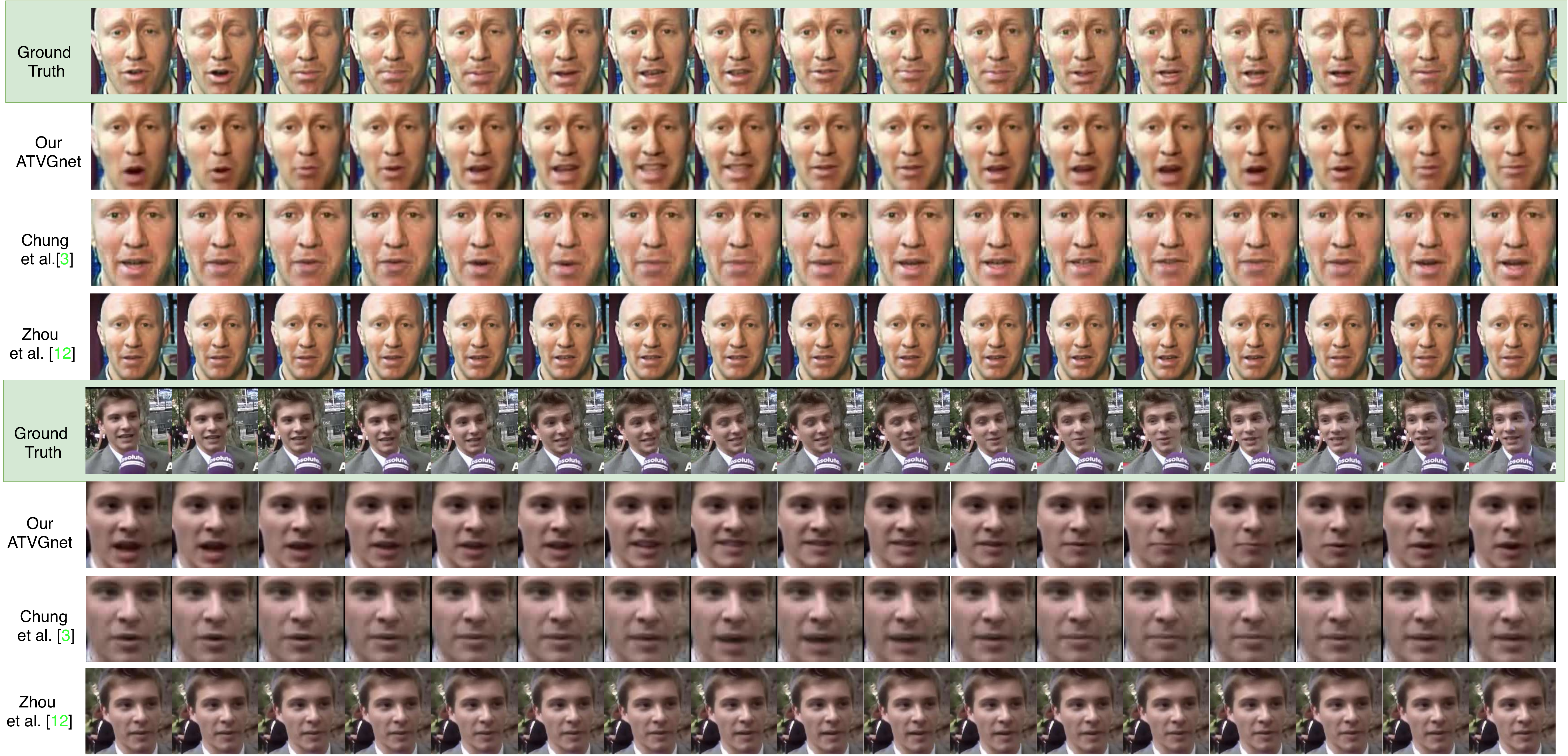}
\caption{Qualitative results produced by ATVGnet, Chung et al.~\cite{ChungJZ17} and Zhou et al.~\cite{zhou2018talking} on samples from LRW and VoxCeleb dataset. We can observe from it that our mouth opening is closer to ground truth compared with the other two methods. It is worthwhile to mention that the second sample is recorded outside with loud background \textbf{noise}.}
\label{fig:visual}
\end{figure*}

Image results are illustrated in Fig.~\ref{fig:result1} and Fig.~\ref{fig:visual}. To evaluate the quality of the synthesized video frames, we compute PSNR and SSIM~\cite{DBLP:journals/tip/WangBSS04}. To evaluate whether the synthesized video contains accurate lip movements that correspond to the input audio, we adopt the evaluation matrix Landmarks Distance (LMD) proposed in~\cite{ChenLMDX18}. We compare our model with other three state-of-the-art methods~\cite{ChenLMDX18,ChungJZ17,DBLP:conf/eccv/WilesKZ18}. All of them are trained on LRW dataset while Chung et al.~\cite{ChungJZ17}
require extra VGG-M network pretrained on VGG Face dataset~\cite{DBLP:conf/bmvc/ParkhiVZ15} and Wilels et al.~\cite{DBLP:conf/eccv/WilesKZ18} need extra MFCC feature extractor pretrained by~\cite{DBLP:conf/accv/ChungZ16a}. The quantitative results are illustrated in Table~\ref{tab:main_tb}. The \text{Baseline} model is a straightforward model without any features (e.g., DMA, MMCRNN, DAL and RD explained in Sec.~\ref{subsec:ablation}) as mentioned in Sec.~\ref{sec:model}. The model \text{ ATVG-ND} has the same network structure as \text{ATVGnet}. But it is trained end-to-end without the decoupled training strategy (see Sec.~\ref{subsec:decouple}). We can find that our ATVGnet achieves the best results both in image quality (SSIM, PSNR) and the correctness of audiovisual synchronization (LMD). 
\begin{figure}[t]
\centering
\includegraphics[width=\linewidth]{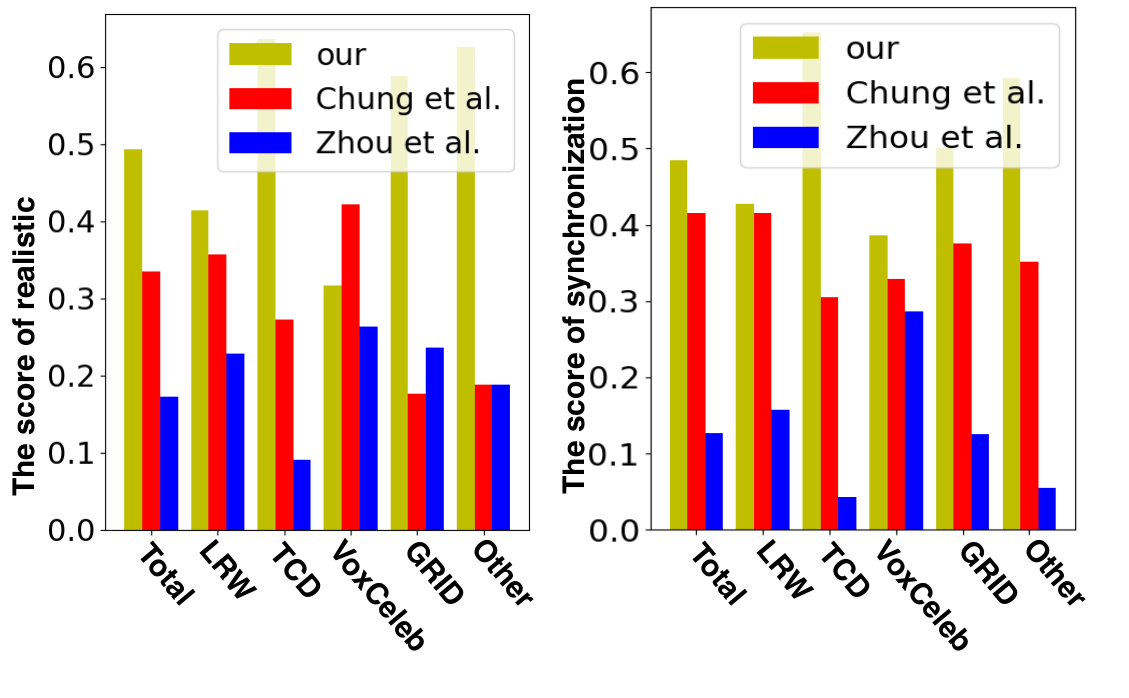}
\caption{Statistics of user studies. The y-axis is the percentage of votes and the x-axis is different data sources (e.g., $total$ means all the video samples, $Other$ means sampled videos from YouTube.) The left histogram is the rating on authenticity. The right histogram is the rating on synchronization between facial movements and audio.}
\label{fig:userstudy}
\end{figure}
\begin{table}[t]
    \centering
  \begin{tabular*}{\linewidth}{  p{1.65cm} p{0.5cm} p{0.6cm} p{0.68cm}|p{0.5cm} p{0.6cm} p{0.6cm} }
      \toprule
      \toprule
Method & \multicolumn{3}{c}{LRW} & \multicolumn{3}{c}{GRID}   \\
      \midrule
& {LMD} &{SSIM}&{PSNR} &{LMD}&{  SSIM}&{ PSNR}  \\
\hline
{Chen~\cite{ChenLMDX18}} &{1.73} &{ 0.73}   &{ 29.65} &{1.59} &{ 0.76}   &{ 29.33}    \\
\hline
{Wiles~\cite{DBLP:conf/eccv/WilesKZ18}} &1.60 &0.75 &29.82 &  1.48 & 0.80 &29.39\\
 \hline
 {Chung~\cite{ChungJZ17}} & { 1.63} &{ 0.77} &{ 29.91} & 1.44& 0.79& 29.87    \\ \hline
   \bottomrule
 { Baseline }   &{{1.71}} &{0.72} &{ 28.95} & { 1.82}     & { 0.77 }  & { 28.78} \\
   \hline
 { ATVG-ND }   &{\bf{1.35}}  &0.78  & { 30.27}&1.34 & { 0.79}  & { 30.51} \\
   \hline
 { ATVGnet }  &1.37 &{ \textbf{0.81}} &{ \bf{30.91}} & {\bf{ 1.29}}     &  {\bf{ 0.83 }}  & {\bf{ 32.15}} \\
      \bottomrule
  \end{tabular*}
  \caption{Quantitative results of different methods on LRW dataset and GRID dataset. Our models mentioned in this table are trained from scratch. We bold each leading score.}
    \label{tab:main_tb}
\end{table}
\subsection{User Studies}
\label{subsec:userstudy}
Our goal is to generate realistic videos based on the audio information. The evaluation in \ref{subsec:quantitative} can only evaluate the quality in a single frame style. To evaluate the performance in a video level, we conduct thoughtful user studies in this section. Human subjects evaluation (see Fig.~\ref{fig:userstudy}) is conducted to investigate the visual qualities of our generated results compared with Chung et al.~\cite{ChungJZ17} and Zhou et al.~\cite{zhou2018talking}. The ground truth videos are selected from different sources: we randomly select samples from the testing set of LRW~\cite{DBLP:conf/accv/ChungZ16a}, VoxCeleb~\cite{Nagrani17}, TCD~\cite{DBLP:journals/tmm/HarteG15}, GRID~\cite{Cooke2006} and real-world samples from YouTube (in total 38 videos). Three methods are evaluated w.r.t. two different criteria: whether participants could regard the generated talking faces as realistic and whether the generated talking faces temporally sync with the corresponding audio. We shuffle all the sample videos and the participants are not aware of the mapping between videos to methods. They are asked to score the images on a scale of 0 (worst) to 10 (best). There are overall 10 participants involved, and the results are summed over persons and video time steps. 

According to the ratings from Fig.~\ref{fig:userstudy}, we can find that our method outperforms other two methods in terms of the extent of synchronization and authenticity. More specifically, our model achieves the best results on all datasets in terms of lip synchronization with audio input. As for image authenticity, our model achieves the highest score the on most of the datasets but slightly lower than Chung et al.~\cite{ChungJZ17} on the VoxCeleb testing set. We attribute this to the audio noise (e.g. background music) in the test samples.
\subsection{Ablation Studies}
\label{subsec:ablation}
\begin{figure}[t]
\centering
\includegraphics[width=1.0\linewidth]{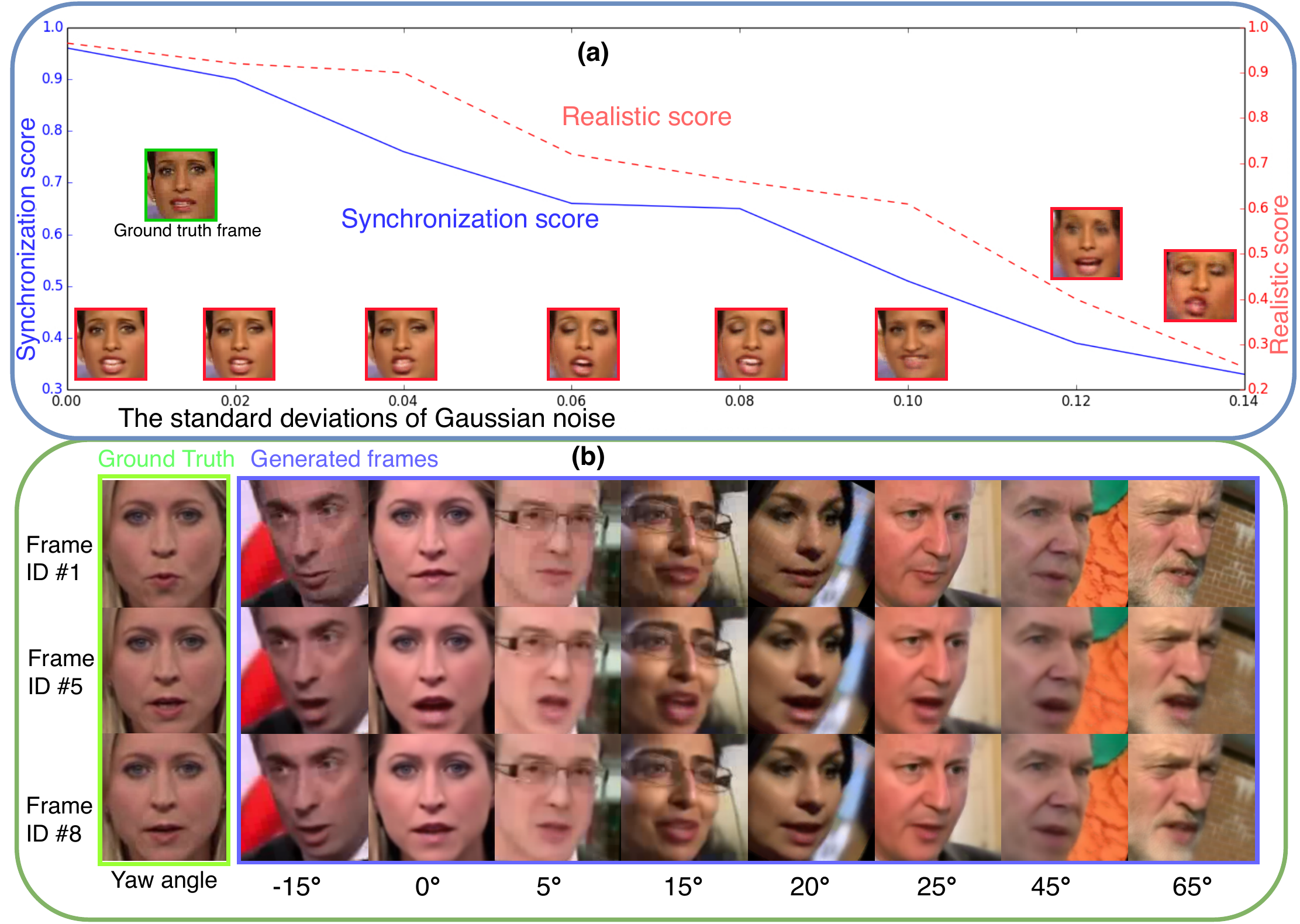}
\caption{The trend of image quality w.r.t. (a) the landmarks (top) and (b) the poses (bottom). Please zoom in on a computer screen.}
\label{fig:noise}
\end{figure}
%
\begin{table}[t] 
      \centering
      \begin{tabular*}{\linewidth}{  p{1.90cm} p{0.5cm} p{0.6cm} p{0.65cm}|p{0.5cm} p{0.5cm} p{0.6cm} }
      \toprule
      \toprule
Method & \multicolumn{3}{c}{LRW} & \multicolumn{3}{c}{GRID}   \\
      \midrule
& { {LMD}} &{  SSIM}&{ PSNR } &{LMD}&{  SSIM}&{ PSNR}  \\
\hline
{ ATVGnet}  & {\bf{ 0.80}}     &  \bf{ 0.86 }  & {\bf{ 33.45}} &{\bf{0.70}} &{ \textbf{0.89}} &{ \bf{33.84}} \\
 
   \hline
    { w/o DMA  }  & { 0.98}     & { 0.83 }  & { 30.22} &{{1.10}} &{0.84} &{ 29.90} \\
   \hline
  {w/o MMCRNN} & {1.03}     & { 0.80 }  & {30.61} &{0.81} &{0.86} &{ 32.68} \\
   \hline
   { w/o DAL }  & { 0.86}     & \bf{ 0.86 }  & {31.35} &{{0.76}} &{0.87} &{ 33.11} \\
   \hline
    { w/o RD }  & { 0.82}     & { 0.84 }  & {32.84} &{0.73} &{0.88} &{ 33.25} \\
   \hline
   { Baseline }  & { 1.27}     & { 0.81 }  & {29.55} &{{1.17}} &{0.80} &{ 29.45} \\
   \hline
     { ATVG-P }  & { 0.90}     & { 0.84 }  & {30.45} &{0.75} &{0.87} &{ 31.78} \\
   \hline
      \bottomrule
  \end{tabular*}
  \caption{Ablation studies on the LRW dataset and the GRID dataset. We remove each feature at a time. We bold the highest scores.}
    \label{tab:ablation}
\end{table}
We conduct ablation experiments to study the contributions of the four components introduced in Sec.~\ref{sec:model}: Dynamic Motion $\&$ Attention (DMA), Multi-Modal-crnn (MMCRNN), Dynamically Adjustable Loss (DAL) and Recreational Discriminator (RD). The ablation studies are conducted on both LRW dataset and GRID dataset. Results are shown in Table~\ref{tab:ablation}. Here we follow the protocols mentioned in Sec.~\ref{subsec:setup}. We test each model using ground truth landmarks rather than fake landmarks generated by AT-net, so that we can eliminate the errors caused by uncorrelated noise and focus on each component.

As shown in Table~\ref{tab:ablation}, each component contributes to the full model. We can find that MMCRNN and DMA are critical to our full model. We attribute this to the better ability of generating smooth transactions between adjacent frames. The \text{ATVG-P} model has the same structure as \text{ATVGnet} but conditioned on the last fake frame $\hat{v}_{t-1}$ rather than the example frame $i_p$ in Eq.~\ref{eq:motion_att} in Sec.~\ref{subsec:attention}. We suppose it could yield better performance. However, the error amplifies quickly through time until it overwhelms the visual information from example frame, which leads to a trivial solution that ${\bm{\alpha}}_{t}= 0_{n \times n}$ and decreases the performance.

We investigate the model performance w.r.t. the generated landmarks accuracy and different pose angles ( see Fig.~\ref{fig:noise}). We add Gaussian noises with different standard deviations to the generated landmarks during inference and conduct user study on the generated videos. The image quality drops (see Fig.~\ref{fig:noise}(a)) if we increase the standard deviation. This phenomenon also indicates that our AT-net can output promising intermediate landmarks. To investigate the pose effects, we test different example images (different pose angles) with the same audio. The results in Fig.~\ref{fig:noise}(b) demonstrate the robustness of our method w.r.t. the different pose angles.

\section{Conclusion and Discussion}
\label{sec:conclusion}

In this paper, we present a cascade talking face video generation approach utilizing facial landmarks as intermediate high-level representations to bridge the gap between two different modalities. We propose a novel Multi-Modal Convolutional-RNN structure, which considers the correlation between adjacent frames in the generation stage. Meanwhile, we propose two novel components: dynamically adjustable loss and regression-based discriminator. In our perspective, these two techniques are general that could be adopted in other tasks (e.g., human body generation and facial expression generation) in the future. Our final model ATVGnet achieves the best performance on several popular datasets in both qualitative and quantitative comparisons. For future work, applying other techniques to enable our network to generate unconscious head movements/expressions could be an interesting topic, which has been bypassed in our current approach.

\noindent \textbf{Acknowledgement.} \quad This work was supported in part by NSF IIS 1741472, IIS 1813709, and the University of Rochester AR/VR Pilot Award. This article solely reflects the opinions and conclusions of its authors and not the funding agents.

{\small
\bibliographystyle{ieee}
\bibliography{egbib}
}

\end{document}